\pdfoutput=1
\documentclass[11pt]{article}

\usepackage[preprint]{acl}

\usepackage{times}
\usepackage{latexsym}
\usepackage{subcaption}
\usepackage{booktabs}
\usepackage[T1]{fontenc}
\usepackage[utf8]{inputenc}

\usepackage{microtype}

\usepackage{inconsolata}

\usepackage{graphicx}

\usepackage{tabularx} 
\usepackage{mathptmx}
\usepackage{amsmath, calc}

\title{Retrofitting Small Multilingual Models for Retrieval: Matching 7B Performance with 300M Parameters}
\author{Lifu Tu, Yingbo Zhou, Semih Yavuz  \\
  Salesforce AI Research \\
  \\}

\begin{document}
\maketitle
\begin{abstract}
Training effective multilingual embedding models is challenging due to linguistic diversity and varying task objectives. While small multilingual models (fewer than 1B parameters) perform well on general multilingual benchmarks, they lag behind larger models in retrieval tasks—the dominant real-world application of embeddings. In this work, we investigate whether smaller models can be specifically optimized for retrieval through a systematic study of publicly available training data and data composition strategies. We analyze the effects of training data scale, negative sampling, and data diversity on multilingual retrieval performance. Our results show that increasing data scale yields diminishing returns, whereas incorporating hard negatives is critical for consistent performance improvements. We further find that task diversity in the training data has a greater impact on retrieval quality than language diversity alone. Guided by these findings, we develop a compact multilingual model with approximately 300M parameters that achieves retrieval performance comparable to, and in some cases surpassing, strong 7B-parameter models.
%Training effective multilingual embedding models presents unique challenges due to the diversity of languages and task objectives. Although small multilingual models (<1 B parameters) perform well on multilingual tasks generally, they consistently lag behind larger models (>1 B) in the most prevalent use case: retrieval. This raises a critical question: Can smaller models be retrofitted specifically for retrieval tasks to enhance their performance? In this work, we investigate key factors that influence the effectiveness of multilingual embeddings, focusing on training data scale, negative sampling strategies, and data diversity. We find that while increasing the scale of training data yields initial performance gains, these improvements quickly plateau—indicating diminishing returns. Incorporating hard negatives proves essential for consistently improving retrieval accuracy. Furthermore, our analysis reveals that task diversity in the training data contributes more significantly to performance than language diversity alone. As a result, we develop a compact (approximately 300M) multilingual model that achieves retrieval performance comparable to—or even surpassing—current strong 7B models.

%\footnote{This draft was completed and patented in May 2025.}
\end{abstract}

\section{Introduction}

%Text embedding models provide dense vector representations to capture semantic information, which can be used find relevant passages or documents from a large corpus can be found given a query. Recently, the significance of high-quality embeddings has been further amplified by their crucial role in retrieval-augmented generation (RAG) systems. Currently, small size multilingual models can works well as shown in MMTEB~\citep{enevoldsen2025mmteb}, which is the a comprehensive test suite encompassing over 100 embedding evaluation tasks across more than 250 languages.  multilingual-e5-large-instruct with only 560 million parameters is the best-performing publicly available model. However, for the majority user case (retrieval task category), small models is much worse (\~ 2 point lower) than large language models (LLMs) with billions of parameters.

Text embedding models~\citep{bert,mt5,izacard2022unsupervised} encode semantic meaning into dense vectors, enabling efficient retrieval of relevant documents from large corpora. Their importance has increased with the rise of retrieval-augmented generation (RAG) systems~\citep{gao2024retrievalaugmentedgenerationlargelanguage}, where retrieval quality directly affects downstream performance. Recent benchmarks such as MMTEB~\citep{enevoldsen2025mmteb}, which evaluates over 100 tasks across more than 250 languages, show that large compact multilingual models can perform well. %For example, multilingual-e5-large-instruct (560M parameters) is the strongest publicly available model~\footnote{This was true at the time of completion of this work; more recently, the Qwen3 Embedding model~\citep{qwen3embedding} reports slightly stronger performance.} overall. 
However, for retrieval—the most common use case—small models still trail billion-parameter LLMs on standard retrieval metrics. %~\footnote{This held at the time of completion; more recent work, such as the Qwen3 Embedding model~\citep{qwen3embedding}, reports strong performance.}. 

In this work, we aim to significantly enhance the retrieval performance of small-scale multilingual embedding models. %—those with around 300 million parameters—bringing them on par with, or even surpassing, the performance of current strong 7B models. 
To accomplish this, we investigate a strategy for retrofitting compact models.% using synthetic multilingual training data derived from the mC4 corpus~\citep{mt5}.
Our approach involves systematically exploring several key factors that influence the effectiveness of multilingual embedding training, including the choice of data sources, the scale of training data, and negative sampling strategies. Through extensive experiments, we uncover several important findings. 
First, we demonstrate that carefully curated synthetic multilingual data, even at small scale (a few thousand examples), yields substantial improvements in retrieval performance. Second, we observe that increasing the amount of training data initially leads to substantial performance gains, but improvements diminish beyond a certain scale, indicating saturation. Third, mining hard negatives with a strong backbone embedding model consistently boosts retrieval accuracy across languages. Finally, our analysis highlights the importance of both task and language diversity in training data, with task diversity having a particularly strong impact on downstream performance.
%Second, we observe that increasing the amount of training data yields substantial performance gains initially, but these gains diminish as the data scale continues to grow, suggesting a point of saturation. Third, we find that mining hard negatives using a strong backbone embedding model consistently enhances retrieval accuracy across languages. Lastly, our analysis highlights the importance of both task diversity and language diversity in training data, with task diversity showing a particularly strong impact on downstream performance.

By prioritizing training strategy over model size, our 300M-parameter model matches or surpasses the retrieval performance of 7B-parameter models on the MMTEB multilingual benchmark, achieving a score of 60.56. Through a systematic study of publicly available data resources and data composition~\footnote{In contrast, for concurrent strong models such as EmbeddingGemma~\citep{vera2025embeddinggemma} and Qwen3-Embedding~\citep{qwen3embedding}, the training data sources and data composition effects are not clearly documented.}, we show that the 3.5-point improvement over the baseline is driven by three key factors: (i) the use of synthetic multilingual query–document pairs generated from the public mC4 corpus~\citep{mt5} with GPT-4o-mini; (ii) the adoption of hard negative mining, which proves more effective than naive data scaling; and (iii) a detailed analysis of data diversity, demonstrating that task diversity has a stronger impact on retrieval performance than simply increasing language coverage.

%By prioritizing training strategy over model size, our 300M-parameter model matches or exceeds the retrieval performance of 7B-parameter models on the MMTEB multilingual benchmark, reaching a score of 60.56. According to the study on public data resource and composition, this 3.5-point gain over the baseline is driven by three key factors: (i) the generation of synthetic multilingual query-document pairs using GPT-4o-mini and the mC4 corpus \citep{mt5}; (ii) the implementation of hard negative mining, which we show to be superior to basic data scaling; and (iii) a systematic analysis of diversity, revealing that task diversity impacts performance more substantially than the mere addition of more languages.

\section{Related work}

\paragraph{Multilingual Embedding Models.} 
Recent work has explored using large language models (LLMs) like XLM-R~\citep{xlmr} and Mistral-7B~\citep{jiang2023mistral7b} as embedding models, demonstrating clear advantages of LLM-based initialization~\citep{wang2023improving, SFRAIResearch2024, snowflake, geminiembedding}. While smaller models, such as multilingual-e5-large-instruct, perform well on benchmarks like MMTEB~\citep{enevoldsen2025mmteb}, their retrieval performance still lags behind larger 7B+ models. This gap limits the adoption of compact models in retrieval-based applications. In this work, we address this limitation by leveraging efficient data. %In this work, we develop a highly effective small multilingual embedding model that matches or surpasses the performance of current 7B models.

%Text embeddings are fundamental to many downstream NLP tasks, such as semantic similarity and information retrieval. Their importance has only increased with the advent of RAG systems, where embedding quality directly affects overall performance. Recently, several studies have leveraged large language models (LLMs) as embedding models~\citep{wang2023improving, SFRAIResearch2024, snowflake, geminiembedding}, initializing from pre-trained architectures like XLM-R~\citep{xlmr}, and Mistral-7B~\citep{jiang2023mistral7b}. These works demonstrate clear benefits of LLM-based initialization.%for embedding quality. Although smaller models—such as multilingual-e5-large-instruct have emerged as the best publicly available options on benchmarks like MMTEB~\citep{enevoldsen2025mmteb}, their retrieval performance still trails significantly behind that of larger 7B+ models. This gap limits the adoption of compact models in retrieval-centric applications, including RAG systems. In this work, we address this limitation by developing a highly effective small multilingual embedding model that rivals or exceeds the performance of current 7B models.

\paragraph{Synthetic Data Generation.} Recent work~\citep{bge_embedding, geminiembedding} has explored using LLMs to generate synthetic data for improving multilingual embeddings, with additional filtering techniques~\citep{thakur-etal-2024-leveraging, geminiembedding} to enhance data quality. In our work, we use clean Wikipedia documents from mC4~\citep{mt5} and GPT-4o-mini to generate synthetic query-document pairs. Our analysis shows that the generated data is high quality and significantly boosts model performance. %Recent work~\citep{bge_embedding, geminiembedding} has explored using large language models (LLMs) to generate synthetic data from documents or passages, aiming to enhance the capabilities of multilingual embedding models. In some cases, additional filtering techniques~\citep{thakur-etal-2024-leveraging, geminiembedding} are employed to improve the quality of the generated data. In our work, we utilize clean Wikipedia documents from mC4~\citep{mt5} as source material and employ a strong generation model (GPT-4o-mini) to create synthetic query-document pairs. Our analysis confirms that the sampled synthetic data is of high quality and contributes meaningfully to model performance.

\paragraph{Small Embedding Models.} %Previously, small embedding models, e.g. ~\citet{bert,bge_embedding,wang2024multilingual} perform not well on retrieval and mainly support short context. Recently, there are also some recent concurrent work, such as EmbeddingGemma~\citep{vera2025embeddinggemma} and Qwen3 Embedding~\citep{qwen3embedding}~\citep to improve this. In our work, we focus more on exploration on data effect (data types, hard negative, language and task diversity) on public models. 
Earlier small embedding models~\citep{bert, bge_embedding, wang2024multilingual} often struggle with retrieval and short-context limitations. Despite recent advancements from concurrent models like EmbeddingGemma~\citep{vera2025embeddinggemma} and Qwen3-Embedding~\citep{qwen3embedding}~\footnote{Data details for the two models are unclear.}, the impact of data resource and composition remains under-explored. In this work, we conduct a rigorous analysis of data effects—including hard negatives and the diversity of tasks and languages—to optimize retrieval performance in public models.

%Recently, several work~\citep{bge_embedding,geminiembedding} use LLM for synthetic data generation given documents or passages to increase the multilingual embedding model ability. Sometimes, filters~\citep{thakur-etal-2024-leveraging,geminiembedding} are used to increase the qualify of these synthetic data. In our this work, we use the clean Wikipedia document from mc4~\citep{mt5} and strong generator (gpt-4o-mini). From the sampled synthetic data, we verify their high qualify.

\section{Method}

\paragraph{Data.} We use existing high-quality and diverse English datasets from multiple task categories, including MS MARCO \citep{nguyen2016ms}, HotpotQA \citep{yang-etal-2018-hotpotqa}, Natural Questions (NQ) \citep{nq}, FEVER \citep{fever}, and AllNLI from SimCSE \citep{gao-etal-2021-simcse}. Additional datasets covering other task types—such as classification, semantic textual similarity (STS), reranking, clustering, and pair classification—are described in Appendix~\ref{data}.

To mitigate the lack of multilingual retrieval data, we construct high-quality synthetic datasets. Specifically, we sample lengthy (100-1000) Wikipedia articles from mC4~\citep{mt5}\footnote{https://huggingface.co/datasets/allenai/c4}, then genera a question given the Wikipedia documents with GPT-4o-mini. The generated question and the sampled article constitute a text pair to the fine-tuning data. We generate synthetic data across 20 languages\footnote{Please see Appendix~\ref{appendix:languages} for the full language list.}. See the discussion on quality of synthetic data in the Appendix~\ref{dataQuality}. %\lifu{adding.}

\paragraph{Training.} Given a batch of query document pairs, we use the contrastive loss~\citep{gao-etal-2021-simcse}: %&\mathrm{L} = 
\begin{align}
\scriptstyle     -\frac{1}{N} \Bigg( \sum_{i=1}^N \log \frac{\exp(\frac{\textit{sim} (h_{q_{i}}, h_{d^+})}{\tau})}{ \exp(\frac{\textit{sim}(h_{q_{i}}, h_{d^+})}{\tau}) + \sum_{j=1}^K \exp( \frac{\textit{sim}(h_{q_{i}}, h_{d_j^-})}{\tau})}  \Bigg)  \nonumber
\end{align}
where $q_i$ is the $i$-th query of the mini batch, $d^+$ is the positive document, and $d_j^-$ is the negative document.  $\tau$ is the temperature term, $N$ is the mini batch size, $K$ is the number of negative documents associated with each query. $h_{q}$ and $h_{d}$ are the representations of the given query $q$ and document $d$, $\textit{sim}$ is the similarity function. Cosine similarity is used in our experiments. We set $\tau=0.02$ in our experiments. More experimental details are shown in the Appendix~\ref{setup}.

\paragraph{Evaluation.} We use MMTEB~\citep{enevoldsen2025mmteb} for multilingual evaluation, which includes over 500 tasks across 250+ languages and multiple domains. We also evaluate on French, Japanese, and other European languages.

%We mainly use MMTEB~\citep{enevoldsen2025mmteb} for the multilingual evaluation , which consists of a large collection of more than 500 distinct tasks covering 250+ languages and multiple domains. Evaluation on French, Japanese and European language are also done.

\section{Results}

 \begin{table*}[ht]
\small
  \centering
  \begin{tabular}{lcccc}
    \toprule
    \textbf{Model} & \textbf{\#Params} & \textbf{Supported Context Length} & \textbf{Emb. Dim.} & \textbf{MMTEB(Retrieval)}\\
    \midrule
     MGTE   & 305M   & 8k  &768 & 56.50    \\
     bge-m3  & 568M  &  8k & 4096 &  55.59    \\
     multilingual-e5-large-instruct    & 560M & 514  & 1024 &  57.12      \\
     snowflake-l   & 568M  & 8k   & 1024 & 58.36      \\
      \midrule
      Our  &  305M  &  8k &  768 &   60.56  \\
        \midrule
        gte-Qwen2-1.5B-instruct & 1.5B & 32k  & 1536 & 60.78 \\
        SFR-Embedding-Mistral & 7B   & 4k & 4096 & 59.44           \\
        gte-Qwen2-7B-instruct & 7B  & 32k  & 3584 & 60.08    \\      
        \bottomrule
  \end{tabular}

  \caption{Performance of models in the MTEB (Multilingual) retrieval tasks. }
  \label{tab:final}
\end{table*}

As shown in prior work~\citep{enevoldsen2025mmteb}, multilingual-e5-large-instruct is the best-performing publicly available model on the MTEB (Multilingual) benchmark. However, it only supports a context length of approximately 500 tokens. Despite its strong performance, its retrieval score is over 3 points lower than that of gte-Qwen2-1.5B-instruct. Furthermore, as shown in Table~\ref{tab:final}, on MTEB (Multilingual) retrieval tasks, the best-performing small multilingual model still lags behind larger models by about 2 points. In this work, we present a compact embedding model (about 300M parameters) that achieves the best performance among small multilingual models on retrieval benchmarks—comparable to or even exceeding that of significantly larger models.

% \begin{table*}
%   \centering
%   \begin{tabular}{lll}
%     \hline
%     \textbf{Model}           & \textbf{natbib command} & \textbf{ACL only command} \\
%     \hline
%     \citep{Gusfield:97}       & \verb|\citep|           &                           \\
%     \citealp{Gusfield:97}     & \verb|\citealp|         &                           \\
%     \citet{Gusfield:97}       & \verb|\citet|           &                           \\
%     \citeyearpar{Gusfield:97} & \verb|\citeyearpar|     &                           \\
%     \citeposs{Gusfield:97}    &                         & \verb|\citeposs|          \\
%     \hline
%   \end{tabular}
%   \caption{\label{citation-guide}
%     Citation commands supported by the style file.
%     The style is based on the natbib package and supports all natbib citation commands.
%     It also supports commands defined in previous ACL style files for compatibility.
%   }
% \end{table*}

\section{Analysis}

\subsection{Data Resource}

In this section, we first examine the usefulness of multilingual synthetic data for retrieval. To explore this, we sample 2,000 examples from three different sources: high-quality English retrieval data from MS MARCO~\citep{nguyen2016ms}, English–French parallel data from OPUS~\citep{tiedemann-2012-parallel}, and our own sampled French synthetic query-document pairs. We evaluate using two backbone embedding models: a multilingual model (MGTE~\citep{zhang-etal-2024-mgte}) and an English-only model (BGE-en-1.5~\citep{bge_embedding}). As shown in Table~\ref{tab:dataresource}, our synthetic data significantly improves multilingual retrieval performance. On the other hand, parallel data contributes little.

We also explore synthetic data generated by other strong publicly multilingual LLMs, such as Qwen/Qwen3-30B-A3B-Instruct-2507~\footnote{See \citet{qwen3technicalreport}}. While similar trends are observed, improvements for non-English languages are smaller than those obtained with GPT-4o-mini. Therefore, in the following experiments, we use GPT-4o-mini exclusively as the generator for multilingual synthetic data.

\begin{table}[h]
  \begin{subtable}{\columnwidth}
  \centering
  \small
  \begin{tabular}{lccc}
    \hline
    \textbf{Setting} & \textbf{all} & \textbf{English} & \textbf{Multilingual}  \\
    \toprule
    MGTE & 56.50 & 49.80 & 67.44 \\
    \midrule
    +English Data     &  58.6 & 52.51 & 68.21       \\
    +parallel data    &  57.24 & 50.68 & 67.54         \\
    +synthetic data   & 58.83  & 52.6 & 68.81      \\
    \bottomrule
  \end{tabular}
  \caption{Multilingual Embedding Model}
  \end{subtable}
  \begin{subtable}{\columnwidth}
  \centering
  \small
  \begin{tabular}{lccc}
    \hline
    \textbf{Setting} & \textbf{all} & \textbf{English} & \textbf{Multilingual}  \\
    \toprule
    BGE-en-1.5 & 38.02 &  45.73 & 25.91 \\ 
    \midrule
    +English Data     & 39.30 & 46.26 & 26.87         \\
    +parallel data    &  37.67 & 45.76 & 23.73          \\
    +synthetic data     &  42.53 & 49.26 & 31.36          \\
    \bottomrule
  \end{tabular}
  \caption{Non-Multilingual Embedding Model}
  \end{subtable}
  \caption{Performance on MTEB (Multilingual) when fine-tuning with different data sources. The column `all', `English', and `Multilingual' refer to the average performance across all retrieval tasks, English-only tasks, and tasks involving non-English languages, respectively.}
  \label{tab:dataresource}
\end{table}

\subsection{Data Scale Effect}

%In the previous subsection, we observed that our synthetic data significantly improves multilingual retrieval performance. This raises an important question: Is it efficient to continue improving retrieval performance by simply adding more synthetic data? To explore this, we sampled French synthetic retrieval data at varying sizes: 1k, 2k, 4k, 8k, and 16k examples. As shown in Figure~\ref{fig:datascale}, multilingual retrieval performance continues to improve with increased data size. However, the performance gains diminish after 4k examples, with each subsequent data doubling yielding smaller improvements compared to the earlier stages. Based on this observation, we choose to generate 4k synthetic samples per language in the following experiments.

\begin{figure}[ht]
\small
  \includegraphics[width=0.9\columnwidth]{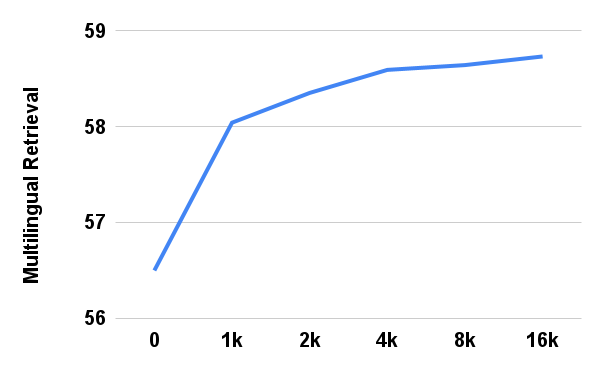}
  \caption{Synthetic Data scale experiments.}
  \label{fig:datascale}
\end{figure}

Building on the observed gains from synthetic data, we examine the effect of data scale by sampling French synthetic retrieval data at sizes from 1k to 16k. As shown in Figure~\ref{fig:datascale}~\footnote{Results of the mean over five runs.}, performance improves with more data but exhibits diminishing returns beyond 4k examples~\footnote{We hypothesize that this is due to the limited domain coverage of the synthetic retrieval data, which is generated from Wikipedia articles in mC4.}. Accordingly, we use 4k synthetic samples per language in subsequent experiments.

% \begin{figure*}[t]
%   \includegraphics[width=0.48\linewidth]{example-image-a} \hfill
%   \includegraphics[width=0.48\linewidth]{example-image-b}
%   \caption {A minimal working example to demonstrate how to place
%     two images side-by-side.}
% \end{figure*}

\subsection{Hard Negatives}

Recent studies~\citep{gecko,geminiembedding} have demonstrated the effectiveness of incorporating hard negatives in training embedding models. In this section, we compare two strategies for selecting hard negatives:
(a) vanilla: neegatives are randomly sampled within each mini-batch.
(b) hard negatives: negatives are mined using the backbone multilingual embedding model.

\begin{figure}[h]
  \small
  \includegraphics[width=0.9\columnwidth]{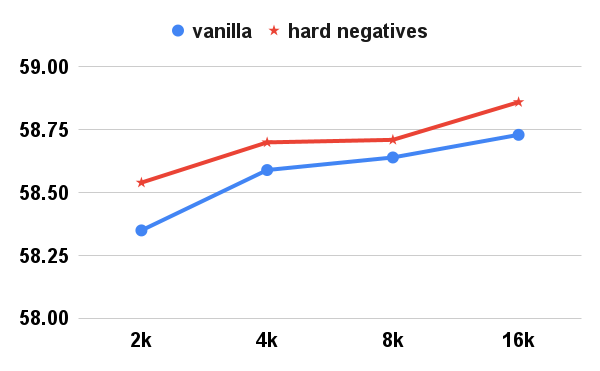}
  \caption{Hard negatives experiments.}
  \label{fig:hardnegative}
\end{figure}

As shown in Figure~\ref{fig:hardnegative}~\footnote{Results of the mean over five runs.}, the hard negative strategy consistently improves multilingual retrieval performance. Based on this observation, we adopt the hard negative mining approach for generating synthetic training samples in our experiments.

\subsection{Task Diversity and Language Diversity}

\begin{table}[ht]
  \centering
  \small
  \begin{tabular}{ccccc}
    \hline
             & \textbf{Mul.} & \textbf{Eur.} & \textbf{Fr} & \textbf{Ja}\\
    
    \hline
        Our (all)  &  60.56   & 59.76 &  68.28  &   72.29               \\
    \hline
     \textbf{En-Diverse}  & 59.60   &  58.22   &  67.56 &  72.26                       \\
     \textbf{Mul-Syn}   &  59.36       &  58.61 & 67.19 & 71.46                        \\
     \textbf{En-Syn + En-Diverse}     &  60.38         &  59.56    & 68.05 & 72.29                   \\

    \hline
  \end{tabular}
  \caption{\label{citation-guide}
    Average retrieval performance across varying task and language diversity. "Mul.", "Eur.", "Fr", and "Ja" denote average scores on MTEB Multilingual, MTEB Europe, French (9 tasks), and Japanese (10 tasks) retrieval benchmarks, respectively%Average retrieval performance across different levels of task and language diversity. "Mul.", "Eur.", "Fr", and "Ja" refer to the average scores on MTEB (Multilingual) retrieval tasks, MTEB (Europe) retrieval tasks, 9 French retrieval tasks, and 10 Japanese retrieval tasks, respectively.
  }
  \label{tab:diversity}
\end{table}

As shown in Table~\ref{tab:diversity}, training our model with all synthetic multilingual data (\textbf{Mul-Syn}) improves MTEB (Multilingual) retrieval performance to 59.36, a substantial gain over the backbone baseline (56.5). To further isolate the effect of data composition, we train a model using only English data but spanning diverse task types (e.g., classification, STS, clustering; \textbf{En-Diverse}). Despite being monolingual, this model performs strongly, underscoring the importance of task diversity.

Combining multilingual synthetic data with English task-diverse data yields our best model, achieving 60.56. Notably, when training with only English synthetic data and English task-diverse data—without any multilingual synthetic data (\textbf{Syn-En + En-Diverse})—performance drops by only 0.2 points on MTEB (Multilingual) retrieval. To better contextualize this difference, we additionally report results on MMTEB (Europe) and on 9 French and 10 Japanese retrieval tasks (Appendix~\ref{appendix: more_retrieval}). Across these languages, performance differences remain minimal, indicating that task diversity contributes more to retrieval gains than language diversity when retrofitting multilingual embedding models.

%As shown in Table~\ref{tab:diversity}, our multilingual embedding model achieves a score of 59.36 on the MTEB (Multilingual) retrieval tasks when trained with all synthetic data (\textbf{Mul-Syn})—a significant improvement over the backbone model baseline (56.5). To further enhance performance, we train a model using all English data (\textbf{En-Diverse}) across a variety of task types (e.g., classification, STS, clustering). This model performs strongly, highlighting the value of diverse task data. By combining all synthetic multilingual data with the English diverse task data, our final model reaches a score of 60.56, improving by another full point. 

%Interestingly, when training only with English synthetic data and English diverse task data (i.e., no multilingual synthetic data) (\textbf{Syn-En + En-Mix}), the resulting model scores just 0.2 points lower on the MTEB (Multilingual) retrieval tasks. To better understand this performance difference, we also report results on MTEB (Europe), as well as on 9 French and 10 Japanese retrieval tasks (listed in the Appendix~\ref{appendix: more_retrieval}). These results show minimal variation across the two models on high-resource languages, suggesting that task diversity contributes more to performance gains than language diversity when continuing to retrofit multilingual embedding models. 

\subsection{Performance Across Other Task Categories}
Table~\ref{tab:mmteb} shows that our models achieve over 1 point higher overall performance and outperform the baseline across most task types. This improvement suggest the effectiveness of our training strategy—particularly the incorporation of diverse task types and synthetic multilingual data—in producing more robust and generalizable multilingual embeddings. The consistent gains across multiple task categories suggest that the model is not overfitting to any single task or language, but is instead learning semantically rich representations that transfer well to a variety of scenarios.

\begin{table}[ht!]
\centering
\small
\begin{tabular}{lc|c}
\toprule
\multicolumn{1}{l}{}  & \textbf{Our} & \textbf{MGTE}  \\ 
\midrule
Mean (Type) & \textbf{52.66} & 51.51  \\
\midrule
- Bitext Mining & 71.61 & 71.79  \\
- Classification & \textbf{58.98} & 57.17  \\
- Clustering & \textbf{46.11} & 44.96   \\
- Inst. Retrieval & -0.85 & -0.74  \\
- Multilabel Class. & \textbf{20.84} & 19.82   \\
- Pair Class. & \textbf{81.01}  & 80.49  \\
- Reranking  & \textbf{61.70} & 60.72   \\
- Retrieval & \textbf{60.56} & 56.50  \\
- STS & \textbf{73.96} & 72.85 \\
\bottomrule
\end{tabular}
\caption{Comparison with the backbone model (MGTE) on the MTEB (Multilingual) benchmark. % for massive multilingual embedding.
}
\label{tab:mmteb}
\end{table}

\section{Conclusion}
In this work, we develop a compact multilingual embedding model (approximately 300M parameters) that achieves retrieval performance comparable to—or even surpassing—that of current strong 7B models. This demonstrates that small models can be effectively used for retrieval, the most critical and practical application of embedding models.

%\subsection{Citations}

% \begin{table*}
%   \centering
%   \begin{tabular}{lll}
%     \hline
%     \textbf{Output}           & \textbf{natbib command} & \textbf{ACL only command} \\
%     \hline
%     \citep{Gusfield:97}       & \verb|\citep|           &                           \\
%     \citealp{Gusfield:97}     & \verb|\citealp|         &                           \\
%     \citet{Gusfield:97}       & \verb|\citet|           &                           \\
%     \citeyearpar{Gusfield:97} & \verb|\citeyearpar|     &                           \\
%     \citeposs{Gusfield:97}    &                         & \verb|\citeposs|          \\
%     \hline
%   \end{tabular}
%   \caption{\label{citation-guide}
%     Citation commands supported by the style file.
%     The style is based on the natbib package and supports all natbib citation commands.
%     It also supports commands defined in previous ACL style files for compatibility.
%   }
% \end{table*}

\section*{Limitations}

In this work, we explore methods to improve the retrieval performance of small multilingual models and demonstrate consistent gains across most task types. However, several limitations remain:
\paragraph{Benchmark} MTEB (Multilingual) covers over 250 languages; however, low-resource languages are primarily represented in task categories such as bitext mining and classification. The evaluation of retrieval performance for these languages remains limited, which constrains our ability to assess the effectiveness of the proposed methods in low-resource settings. This limitation may also explain why the improvement from multilingual synthetic data—compared to English-only synthetic data—is relatively small, as shown in Table~\ref{tab:diversity}. Because of the limitation of the existing benchmark, it is hard to evluate performance on very low-resource languages~\footnote{Although the used benchmark covers over 250 languages, retrieval performance on low-resource languages is already high, leaving limited room for further improvement.}.

\paragraph{Efficiency} In this work, we show that multilingual synthetic data effectively improves retrieval performance, but its benefits eventually plateau, indicating a limitation when relying on retrieval data on few domains alone. We further find that incorporating diverse non-retrieval task data leads to additional gains. Scaling the joint use of diverse non-retrieval task data and multilingual synthetic retrieval data is a promising direction for future research.

\bibliography{custom}

\appendix

\section{Details on Synthetic Data Generation}
\subsection{Languages for Synthetic Data} \label{appendix:languages}
List of  language ISO-639 code (\url{https://cloud.google.com/translate/docs/languages}): en, fr, es de,zh, it, ja, ko, fa, hi, id, ar, bn, fi, sw, te, th, jv, ms, sq.

\subsection{Template for Synthetic Data Generation}.
Give a sampled Wikipedia documents \texttt{\{d\}} and target language \texttt{\{lang\}}, the following template is used to generate a query: 

\begin{quote}
    \texttt{You are a curious AI assistant, please generate one specific and valuable question based on the following text. If the text is not \{lang\}, please reply with Non-\{lang\}. The generated question should revolve around the core content of this text, and avoid using pronouns (e.g., 'this'). Note that you should generate only one question, without including additional content:\\
    \{d\}
    }
\end{quote}

\section{Models}
Below is a list of models discussed in this work:
\begin{itemize}
    \item MGTE~\citep{zhang-etal-2024-mgte}\footnote{Alibaba-NLP/gte-multilingual-base}
    \item BGE-en-1.5~\citep{bge_embedding}\footnote{BAAI/bge-large-en-v1.5}
    \item bge-m3~\citep{chen-etal-2024-m3}\footnote{BAAI/bge-m3}
    \item multilingual-e5-large-instruct~\citep{wang2024multilingual}\footnote{intfloat/multilingual-e5-large-instruct}
    \item snowflake-l~\citep{snowflake}\footnote{Snowflake/snowflake-arctic-embed-l-v2.0} 
    \item gte-Qwen2-1.5B-instruct~\citep{li2023towards}\footnote{Alibaba-NLP/gte-Qwen2-1.5B-instruct} 
    \item SFR-Embedding-Mistral~\citep{SFRAIResearch2024}\footnote{Salesforce/SFR-Embedding-Mistral}
    \item gte-Qwen2-7B-instruct~\citep{li2023towards}\footnote{Alibaba-NLP/gte-Qwen2-7B-instruct}
\end{itemize}

In this work, the AI assistant ChatGPT is used solely for writing and language polishing.

\section{Additional Data Details}
\label{data}
The following are additional English data samples from the training split of MTEB, excluding data used in the official MTEB evaluation.
\begin{itemize}
    \item Classification: AmazonPolarity, AmazonReviews, Imdb, MTOPDomainClassification,AmazonReviewsClassification
    \item STS:  STS12, STS22, STSBenchmark
    \item Reranking: SciDocsRR and StackOverflowDupQuestions
    \item Clustering: arXIV, bioRxiv, medRxiv
    \item Pair Classification: SprintDuplicateQuestions, TwitterURLCorpus
\end{itemize}

\section{Quality of synthetic data}
\label{dataQuality}
In order to avoid "model collapse" when training on synthetic data, we take several ways to improve the quality of the synthetic data:
\begin{itemize}
    \item  High-quality multilingual Wikipedia documents (e.g., mC4; \citealp{mt5}) are used as clean source material for synthetic data generation.  %High quality multilingual Wikipedia documents ( mC4~\citep{mt5} are clean source material) when is used for synthetic data generation.
    \item For each ⟨question, document⟩ pair, only the question is generated by a large language model, while the document is drawn from existing sources. We further restrict the document length to between 100 and 1,000 tokens.
    \item To assess question quality, we randomly sampled 50 questions for each language and manually evaluated them. Surprisingly, most questions are well-formed and relevant to the given documents.
\end{itemize}

\section{Implementation Details}
\label{setup}
We use a batch size of 1024 and a learning rate of 1e-5, with 100 warm-up steps followed by linear decay. The maximum sequence lengths are set to 256 for queries and 512 for documents. LoRA adapters with a rank of $r$ = 8 are applied. Each query-document pair is trained with 7 hard negatives, all drawn from the same task.

\section{Additional Retrieval Tasks}
\label{appendix: more_retrieval}
\paragraph{9 French retrieval tasks:} French subset of the following 9 tasks: AlloprofRetrieval, BSARDRetrieval, FQuADRetrieval, SyntecRetrieval, BelebeleRetrieval, MIRACLRetrievalHardNegatives, PublicHealthQA, StatcanDialogueDatasetRetrieval, WebFAQRetrieval

\paragraph{10 Japanese retrieval tasks: } Japanese subset of the following 10 tasks: JaGovFaqsRetrieval, JaqketRetrieval, JaQuADRetrieval, NLPJournalAbsIntroRetrieval, NLPJournalTitleAbsRetrieval, NLPJournalTitleIntroRetrieval, BelebeleRetrieval, MIRACLRetrievalHardNegatives, MrTidyRetrieval, WebFAQRetrieval.

\end{document}